\documentclass[11pt]{article}

\usepackage[preprint]{acl}

\makeatletter
\IfFormatAtLeastTF{2025-06-01}{
  \AddToHook{build/column/before}{
    \protected@write\@auxout{}{
      \string\def\string\@LN@column{\if@firstcolumn2\else1\fi}
    }
  }
}{}
\makeatother

\usepackage{times}
\usepackage{latexsym}

\usepackage[T1]{fontenc}

\usepackage[utf8]{inputenc}

\usepackage{microtype}

\usepackage{inconsolata}

\usepackage{graphicx}

\usepackage{microtype}
\usepackage{graphicx}
\usepackage{subfigure}
\usepackage{booktabs}

\usepackage{amsmath}
\usepackage{amssymb}
\usepackage{mathtools}
\usepackage{amsthm}

\usepackage{booktabs}
\usepackage{threeparttable}
\usepackage{siunitx}
\usepackage{longtable}
\usepackage{graphicx}
\usepackage{tabularx}
\usepackage{array}
\usepackage{multirow}
\usepackage{color,colortbl}
\usepackage{makecell}
\usepackage{rotating}
\usepackage{xspace}
\usepackage{bbm}
\usepackage{capt-of}

\newcommand{\custompara}[1]{{\vspace{0.4mm}\noindent\textbf{#1}\xspace}}

\usepackage[capitalize,noabbrev]{cleveref}

\usepackage{tcolorbox}
\tcbuselibrary{skins}
\newtcolorbox{prompt}[1]{
  enhanced,
  left=2mm, right=2mm,
  top=1mm, bottom=1mm,
  boxsep=0mm,
  boxrule=0.4pt,
  rounded corners,
  title=#1,
  fontupper=\scriptsize\linespread{0.88}\selectfont,
}
\newenvironment{lmttfont}{\fontfamily{lmtt}\selectfont}{\par}

\theoremstyle{plain}

\theoremstyle{definition}

\theoremstyle{remark}

\usepackage{xcolor}

\title{Text-to-Stage: Spatial Layouts from Long-form Narratives}

\author{
  Jefferson Hernandez\textsuperscript{1},
  Swarnadeep Saha\textsuperscript{2},
  Chenxi Whitehouse\textsuperscript{2},
  Sanjeel Parekh\textsuperscript{3},
  Calvin Murdock\textsuperscript{3}, \\
  \textbf{Yuliang Li\textsuperscript{3}},
  \textbf{W. Owen Brimijoin\textsuperscript{3}},
  \textbf{Vamsi Krishna Ithapu\textsuperscript{3}},
  \textbf{Ishwarya Ananthabhotla\textsuperscript{3}} \\
  \textsuperscript{1}Rice University \quad
  \textsuperscript{2}Meta FAIR \quad
  \textsuperscript{3}Meta Reality Labs Research \\
}

\begin{document}
\maketitle

\begin{abstract}

In this work, we probe the ability of a language model to demonstrate spatial reasoning from unstructured text, mimicking human capabilities and automating a process that benefits many downstream media applications.  
Concretely, we study the narrative-to-play task: inferring stage-play layouts (scenes, speaker positions, movements, and room types) from text that lacks explicit spatial, positional, or relational cues.  We then introduce a dramaturgy-inspired deterministic evaluation suite and, finally, a training and inference recipe that combines rejection SFT using Best-of-N sampling with RL from verifiable rewards via GRPO.  Experiments on a text-only corpus of classical English literature demonstrate improvements over vanilla models across multiple metrics (character attribution,  spatial plausibility, and movement economy), as well as alignment with an LLM-as-a-judge and subjective human preferences.
\end{abstract}  
\section{Introduction}
\label{sec:intro}

\begin{figure}[h!]
    \centering
    \includegraphics[width=0.8\linewidth]{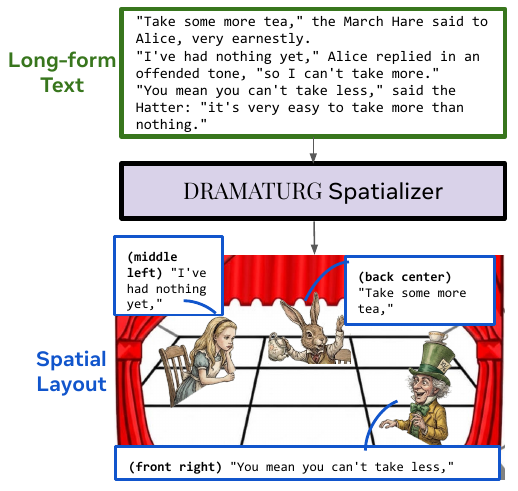}
\caption{We propose a \textsc{Dramaturg} Spatializer that transforms long-form narrative into structured stage layouts. The model assigns dialogue and characters to canonical positions on a discrete grid---depth (\textit{front, middle, back}) and lateral placement (\textit{left, center, right})---with temporal blocking to produce plausible staging from text. This automates a ``text-to-play'' process for applications such as game previsualization and spatial audiobooks.}
\label{fig:teaser}
\vspace{-0.6em}
\end{figure}
 
\begin{figure*}[ht!]
    \centering
    \includegraphics[width=\linewidth]{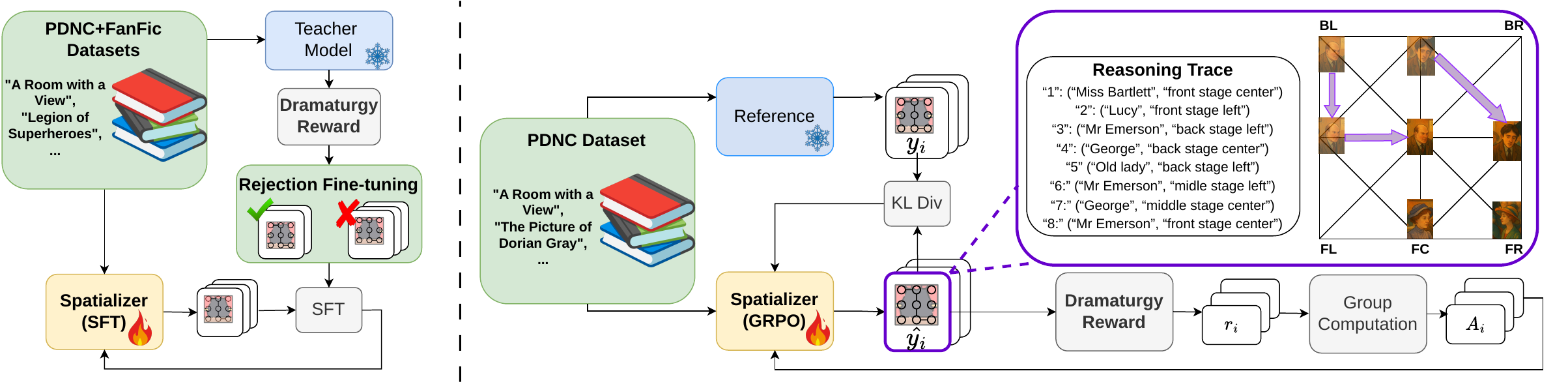}
\caption{\textbf{Overview of our Spatializer training pipeline.}
\textbf{Left:} A strong teacher LLM produces pseudo stage-play annotations via Best-of-$N$ sampling and \emph{rejection SFT} under our deterministic dramaturgy evaluator.
\textbf{Right:} The resulting \emph{Spatializer-SFT} is post-trained with \emph{RL from verifiable rewards} (GRPO): for each passage, we sample candidate spatializations, score them with the deterministic evaluator, compute group-relative advantages (with KL regularization to a frozen reference model), and update the policy to obtain \emph{Spatializer-GRPO}. The model outputs a reasoning trace (e.g., per-quote placement plan) and the corresponding stage-grid layout.}
\label{fig:system}
\vspace{-0.8em}
\end{figure*}

Humans can infer spatial configurations of speakers and characters while reading, even when explicit positional cues are sparse; this allows for better comprehension of and reasoning about the text \citep{zwaan1998situation, MORROW1987165}. Directors and media creators perform a similar inference process when turning scripts or prose into blocking and previsualization \citep{hodge2015play,bloom2001thinking}. Here, we ask whether large language models (LLMs) can produce comparable spatial layouts from long-form text.

We study the \emph{text-to-stage} task: given a contiguous narrative window, the model outputs a structured stage-play representation that (i) segments scenes, (ii) attributes quotes and resolves aliases, (iii) places characters on a discrete stage grid, and (iv) annotates entrances and movements (\autoref{fig:teaser}). High-quality staging requires global and temporal reasoning beyond ``who spoke'': proxemic distance and depth salience \citep{hall1966hidden}, left--right balance and visual weight \citep{mcmanus2011arnheim}, stable composition with clear primary/secondary focus \citep{block2020visual,kress2020reading}, and movement economy \citep{hodge2015play,bloom2001thinking}.

This setting is challenging for LLMs because inputs are long and underspecified, and evaluation is inherently multi-solution. Narrative text provides sparse spatial cues, so the model must jointly infer geometry, salience, and continuity over long horizons \citep{yamada2023evaluating,mirzaee2021spartqa,shi2022stepgame,wang2024picture}. To make progress measurable, we define deterministic constraints and component-wise scores grounded in core dramaturgical and visual-composition principles \citep{hall1966hidden,mcmanus2011arnheim,block2020visual,kress2020reading,hodge2015play,bloom2001thinking}. This yields a verifiable objective that also supports selection and post-training with verifiable feedback \citep{cobbe2021training,wang2022self,shao2024deepseekmath,wen2025reinforcement}.

Concretely, we formulate a new task and JSON schema for implicit book-to-play spatialization. We build a deterministic evaluator for validity, spatial plausibility, movement coherence, and scene transitions, and use it for both Best-of-$N$ selection and RL from verifiable rewards. Starting from rejection-SFT on teacher-generated data, GRPO further improves plausibility and movement economy; for example, our GRPO spatializer reaches an $88.5\%$ macro deterministic score versus $85.1\%$ for SFT (Table~\ref{tab:leaderboard}). We evaluate prompt-only and trained models with (i) the deterministic suite, (ii) LLM-as-judge scoring, and (iii) a human preference study on binaural spatialized audiobook clips.

Our contributions are:
\vspace{-0.1in}
\begin{itemize}
  \item[(C1)] A new task formulation for implicit spatial reasoning from long-form narrative, mapping book passages to stage-play JSON with scenes, speakers, positions, and movements.
  \item[(C2)] A deterministic dramaturgical evaluation suite that instantiates theatre and visual-composition principles (proxemics, balance/focus, movement economy, and scene transitions) as verifiable constraints and component scores.
  \item[(C3)] A training and inference recipe for dramaturgy-aware spatialization that combines rejection SFT, Best-of-$N$ selection, and RL from verifiable rewards with GRPO, improving spatial plausibility and movement coherence on our task.
\end{itemize}

\section{Related Work}
\label{sec:related}

\vspace{-0.1in}
\custompara{From narrative to scripts.}
Closest to our setting are LLM tools for screenplay or theatre-script drafting (e.g., Dramatron, HoLLMwood), which convert prose into dialogue and scene summaries \citep{mirowski2023co,chen2024hollmwood}. These systems target \emph{textual} authoring, not explicit spatial placement and movement planning that preserves source text. We instead predict structured stage blocking directly from narrative under verifiable constraints.

\custompara{Generative models for layout composition.}
LLM-based visual planners map natural-language briefs to structured placements. Text2Scene, LayoutGPT, LayoutPrompter, and PosterLLaMA serialize constraints so sequence models can generate and rank layouts \citep{tan2019text2scene,feng2023layoutgpt,lin2023layoutprompter,seol2024posterllama}. In parallel, VAE-, diffusion-, and flow-based methods (CanvasVAE, LayoutDM, LayoutDiffusion, LDGM, DiffDocLayout) produce controllable 2D layouts under partial constraints \citep{yamaguchi2021canvasvae,chai2023layoutdm,zheng2023layoutdiffusion,hui2023unifying,he2023diffusion}. We follow this line by using explicit, scored dramaturgy constraints for reliability.

\custompara{Agentic and RL-driven layout with verifiable feedback.}
Verifier-based selection improves reasoning by sampling candidates and choosing those that satisfy a checker \citep{cobbe2021training,wang2022self}. RL with verifiable rewards (RLVR) directly optimizes checkable objectives; GRPO is a compute-efficient variant used in recent reasoning models \citep{shao2024deepseekmath,wen2025reinforcement}. Our method similarly optimizes verifiable theatre-inspired signals (proxemics, composition, and movement economy) over temporal dialogue.

\custompara{Narrative-to-scene grounding in 2D/3D.}
Text-to-scene pipelines map language to 2D/3D layouts through structured intermediates, retrieval, and procedural assembly \citep{tan2019text2scene,johnson2018image}. Recent language-to-3D systems (SceneTeller, GALA3D, Ctrl-Room, LLplace) also plan layouts before rendering or optimization \citep{ocal2024sceneteller,yang2024llplace,zhou2402gala3d,fang2025ctrl}. Our setting differs because the ``objects'' are dialogue speakers on a rule-governed stage grid, but these works support LLMs as high-level spatial planners.

\custompara{Benchmarks for spatial reasoning in LLMs.}
Many spatial-reasoning benchmarks evaluate geometric competence through text QA or symbolic outputs (e.g., SPARTQA, StepGame) \citep{mirzaee2021spartqa,shi2022stepgame,yamada2023evaluating,wang2024picture}. Our narrative-to-plays suite complements them by evaluating long-form \emph{text-to-layout} plausibility and movement economy with deterministic metrics and LLM-as-judge preferences \citep{zheng2023judging}.

\section{Methodology}
\label{sec:method}
\vspace{-0.1in}

We study \emph{text-to-stage} spatialization for long-form narrative content. Given a contiguous window of book text \(\mathbf{x}\) (chapter segment, scene, or paragraph group), our system predicts a structured stage-play plan \(\mathbf{y}\) (~\autoref{fig:system}), with the components described in \autoref{sec:method:layout}. Concretely, our goal is to train an LLM spatializer \(\pi_\theta(\mathbf{y}\mid\mathbf{x})\) that, given a dramaturgy prompt (see \autoref{sec:prompts}) and a book text \(\mathbf{x}\), generates a reasoning trace $\mathbf{t}$ and the plan \(\mathbf{y}\). In particular, we propose a synthetic data generation and post-training recipe that optimizes the LLM via Supervised Fine-tuning (SFT), followed by Reinforcement Learning with Verifiable Rewards (RLVR).

\subsection{Stage-play Layout}
\label{sec:method:layout}
\vspace{-0.1in}

The spatializer emits a temporally ordered JSON with fields for (a) scene headers and boundaries, (b) a per-scene cast list with canonical names, and (c) a quote-level table of speaker labels and placements. Placements are defined on a discrete stage grid \(\mathcal{G}\) (default \(3{\times}3\)), interpreted as \(\{\text{front},\text{middle},\text{back}\}\times\{\text{left},\text{center},\text{right}\}\), where depth approximates downstage/upstage salience and lateral position supports comprehensible blocking. To keep the representation verifiable, we enforce simple structural conventions in the schema, such as fixed cell labels and required per-quote fields. \autoref{fig:system} (right) provides a qualitative illustration of the model output.

\vspace{-0.1in}
\subsection{Dramaturgy-inspired Verifiable Reward}
\label{sec:method:deterministic-reward}
\vspace{-0.1in}

Staging from narrative is inherently underdetermined: many distinct layouts can be plausible for the same input text. To make quality measurable and directly optimizable; we design a deterministic evaluator \(R(\mathbf{x},\mathbf{y})\in[0,1]\) derived from widely taught principles in dramaturgy, directing, and visual composition: proxemics and depth salience \citep{hall1966hidden}, balance and visual weight \citep{mcmanus2011arnheim,block2020visual}, clear primary/secondary focus and stable composition \citep{block2020visual,hodge2015play}, and economical movement \citep{hodge2015play,bloom2001thinking}. The evaluator decomposes dramaturgical competence into five components: \textbf{(1) structural validity}, \textbf{(2) text grounding}, \textbf{(3) spatial composition}, \textbf{(4) movement coherence}, and \textbf{(5) scene structure}. Each component produces a normalized score in \([0,1]\), and we combine them with a convex linear combination (invalid JSON receives zero reward). We use uniform weights by default; the exact subchecks, thresholds, and any per-component scaling details are specified in \autoref{sec:app:deterministic-reward}.

\custompara{(1) Structural validity.}
The first failure mode for long structured outputs is malformed JSON or schema drift. We therefore condition reward by schema validity and assign explicit credit only when required fields are present and correctly typed. This makes evaluation zero-noise and ensures that downstream selection and RL do not “reward hack” by emitting unparseable outputs.

\custompara{(2) Text grounding.}
A layout is only meaningful if it is anchored to the dialogue. We score whether the predicted speaker label matches the quote-level attribution and whether mentions/aliases are normalized to canonical character identities. This component prevents the model from improving “layout aesthetics” by hallucinating speakers or merging distinct entities.

\custompara{(3) Spatial composition.}
We operationalize a small set of composition rules that make blocking legible. Depth acts as a proxy for prominence: primary interlocutors should tend to occupy downstage positions, reflecting salience and conversational closeness, while crowding of individual cells is discouraged \citep{hall1966hidden}. We additionally score left--right balance (avoiding excessive visual weight on one side) and encourage distinct placements among the most active characters to maintain clear primaries and comprehensible negative space \citep{mcmanus2011arnheim,block2020visual}. Finally, we reward stable, non-degenerate staging patterns (e.g., avoiding repeated colinear “stacking” of primary characters) to preserve focus across extended exchanges \citep{block2020visual,hodge2015play}.

\custompara{(4) Movement coherence.}
Directing practice emphasizes that movement should be legible, motivated, and used sparingly: constant repositioning without narrative cause breaks continuity and draws attention away from the dialogue \citep{hodge2015play,bloom2001thinking}. We therefore reward small, local steps; prefer lateral adjustments over frequent depth oscillations; and penalize “thrashing” patterns (e.g., repeated front--back flips over consecutive turns). This yields dense feedback that targets a common LLM spatializer failure mode: gratuitous motion inserted to “look dynamic.”

\custompara{(5) Scene structure and transitions.}
We score whether predicted scene boundaries coincide with coherent shifts in location/time/dramatic focus. We also include transition logic that encourages continuity of blocking across scene breaks (e.g., consistent carry-over placements for continuing characters) and legible entrances/exits (e.g., new entrants appear upstage unless otherwise justified), reflecting standard blocking conventions \citep{hodge2015play,bloom2001thinking}.

\subsection{SFT with Synthetic Examples}
\label{sec:method:synthetic-data}
\vspace{-0.1in}

In place of costly human annotations, we use the deterministic evaluator to generate high-quality synthetic examples as supervision. We construct a new synthetic dataset (specifics in Section~\ref{sec:exp:data}) and, for each unlabeled text window \(\mathbf{x}\), sample \(N\) candidate layouts \(\{\mathbf{y}_i\}_{i=1}^N\) from a strong teacher model. The teacher model is prompted with our \textit{Stage-play Generation Prompt}, given in \autoref{sec:prompts}.  We score each candidate with \(R(\mathbf{x},\mathbf{y}_i)\) and keep the highest-scoring output, optionally filtering out candidates below a minimum quality threshold. This is a practical, verifier-guided variant of self-consistency selection for structured generation \citep{cobbe2021training,wang2022self}. We then perform supervised fine-tuning 
on the accepted synthetic pairs \((\mathbf{x},\mathbf{y}^\star)\). Because the synthetic targets are selected (and optionally rejected) under the same deterministic criteria used for evaluation, this procedure acts as \emph{rejection SFT}: the student learns to imitate outputs that already satisfy dramaturgy-inspired constraints.

\subsection{Post-training with GRPO}
\label{sec:method:training}

The SFT approach outlined in \autoref{sec:method:synthetic-data} is inherently limited in that model can only learn from the highest-scoring examples. Therefore, we further improve the SFT-trained Spatializer by directly optimizing the deterministic rewards using Reinforcement Learning from Verifiable Rewards (RLVR) with Group Relative Policy Optimization (GRPO) \citep{shao2024deepseekmath,wen2025reinforcement}. For each input \(\mathbf{x}\), we sample a group of trajectories \(G\), compute rewards \(r_i=R(\mathbf{x},\mathbf{y}_i)\), and form the group-normalized advantage $A_i$; we then maximize the clipped KL-regularized objective against the output of the reference model, following \citep{shao2024deepseekmath,wen2025reinforcement}.  This critic-free, group-baseline estimator encourages trajectories that improve the deterministic dramaturgical score while the KL penalty stabilizes training by constraining drift from the reference policy.

\section{Experiment Settings}
\label{sec:experiments}

\subsection{Datasets}
\label{sec:exp:data}

\textbf{PDNC}
We construct our primary evaluation dataset from the Project Dialogism Novel Corpus (PDNC)~\cite{vishnubhotla2022project}, which provides manual annotations for quotation attribution and alias resolution across classic English novels. We follow the original PDNC split (PDNC1 training, PDNC2 validation) and the preprocessing of \cite{michel2025evaluating}: each chapter is chunked into contiguous \(4096\)-token windows (roughly half a chapter), yielding \(\sim 1\text{k}\) instances total (\(\sim 800\) train, \(\sim 200\) validation).
PDNC labels supervise \emph{quotation attribution} and \emph{alias resolution} directly. For the remaining stage-play JSON fields (scene segmentation, per-quote positions, entrances/movements, and scene metadata), 
we use a strong teacher LLM (``Llama-4-Maverick''): for each instance, we sample \(64\) candidates (\emph{Best-of-64}), reject invalid/inconsistent JSON (e.g., out-of-bounds grid positions), and keep the highest-quality candidate under the same deterministic checks used by our evaluator (\autoref{sec:method:deterministic-reward}).

\begin{table*}[ht!]
  \centering
  \small
  \newcommand{\grayfont}{\color{gray}}
  \renewcommand{\arraystretch}{1.3}
  \setlength{\tabcolsep}{5pt}
  \begin{tabular}{l c c c c c c c}
    \toprule
    \multirow{2}{*}{\textbf{Model}} & \multicolumn{6}{c}{\textbf{Deterministic suite components}} & \multirow{2}{*}{\textbf{AVG}} \\
    \cmidrule(lr){2-7}
    & \textbf{\shortstack{Quote\\Attr.}} 
    & \textbf{\shortstack{Alias\\Res.}} 
    & \textbf{\shortstack{Stage\\Pos.}} 
    & \textbf{\shortstack{Char.\\Pos.}} 
    & \textbf{\shortstack{Move.\\Coh.}} 
    & \textbf{\shortstack{Scene\\Trans.}} 
    & \\
    \midrule
    \multicolumn{8}{l}{\textit{\grayfont Proprietary models}} \\
    \midrule
    \grayfont Claude Sonnet 3.5      & \grayfont \textbf{96.0} & \grayfont 97.0 & \grayfont 87.0 & \grayfont 85.0 & \grayfont 77.0 & \grayfont 51.0 & \grayfont 82.0 \\
    \grayfont GPT-4.1                & \grayfont 94.0          & \grayfont 99.0 & \grayfont 65.0 & \grayfont 84.0 & \grayfont 90.0 & \grayfont 51.0 & \grayfont 81.0 \\
    \grayfont GPT-4o                 & \grayfont 95.0          & \grayfont 99.0 & \grayfont 68.0 & \grayfont 83.0 & \grayfont 84.0 & \grayfont 51.0 & \grayfont 80.0 \\
    \grayfont Claude Sonnet 4        & \grayfont \textbf{96.0} & \grayfont 97.0 & \grayfont 85.0 & \grayfont 83.0 & \grayfont 69.0 & \grayfont 50.0 & \grayfont 80.0 \\
    \grayfont Gemini-2.0-flash       & \grayfont 93.0          & \grayfont 96.0 & \grayfont 68.0 & \grayfont 81.0 & \grayfont 86.0 & \grayfont 51.0 & \grayfont 79.0 \\
    \grayfont Claude Sonnet 3.7      & \grayfont 94.0          & \grayfont 97.0 & \grayfont 88.0 & \grayfont 83.0 & \grayfont 64.0 & \grayfont 49.0 & \grayfont 79.0 \\
    \midrule
    \multicolumn{8}{l}{\textit{\grayfont Open-source models}} \\
    \midrule
    \grayfont Llama-4-Scout          & \grayfont 92.0          & \grayfont 99.0          & \grayfont 71.1 & \grayfont 86.8 & \grayfont 90.5 & \grayfont \textbf{61.4} & \grayfont 83.4 \\
    \grayfont Llama-3.3-70B-it       & \grayfont 92.5          & \grayfont 98.9          & \grayfont 71.6 & \grayfont 86.9 & \grayfont 90.2 & \grayfont 60.4          & \grayfont 83.4 \\
    \grayfont Llama-4-Maverick       & \grayfont \textbf{96.0} & \grayfont \textbf{99.9} & \grayfont 74.3 & \grayfont 85.6 & \grayfont 87.4 & \grayfont 55.7          & \grayfont 83.1 \\
    \grayfont Llama-3.3-8B-it        & \grayfont 86.8          & \grayfont 96.8          & \grayfont 82.2 & \grayfont 84.3 & \grayfont 83.7 & \grayfont 56.2          & \grayfont 81.7 \\
    \grayfont Gemma-3-27b-it         & \grayfont 89.1          & \grayfont 98.1          & \grayfont 80.3 & \grayfont 73.7 & \grayfont 78.3 & \grayfont 60.4          & \grayfont 80.0 \\
    \grayfont Qwen3-8b               & \grayfont 83.8          & \grayfont 95.1          & \grayfont 68.1 & \grayfont 79.7 & \grayfont 91.2 & \grayfont 51.1          & \grayfont 78.2 \\
    \grayfont Llama-3.1-8B-it        & \grayfont 27.0          & \grayfont 31.6          & \grayfont 23.0 & \grayfont 27.6 & \grayfont 29.3 & \grayfont 15.1          & \grayfont 25.6 \\
    \midrule
    \multicolumn{8}{l}{\textit{Our models}} \\
    \midrule
    Spatializer-SFT                  & 92.8 & 98.2 & 77.7          & 87.2          & 96.1          & 57.0 & 84.8 \\
    \rowcolor{blue!5}
    Spatializer-GRPO                 & 91.8 & 99.1 & \textbf{92.1} & \textbf{92.5} & \textbf{98.2} & 57.4 & \textbf{88.5} \\
    \bottomrule
  \end{tabular}
  \vspace{-0.15em}
  \caption{Leaderboard across models. Metrics include \textbf{Quote Attr.} (Quote Attribution), \textbf{Alias Res.} (Alias Resolution), \textbf{Stage Pos.} (Stage Positioning Validity), \textbf{Char. Pos.} (Character Positioning), \textbf{Move. Coh.} (Movement Coherence), and \textbf{Scene Trans.} (Scene Transitions). Top scores are bolded.}
  \label{tab:leaderboard}
\vspace{-0.6em}
\end{table*}

\begin{figure*}[ht!]
    \centering
    \subfigure[\small Distribution of scores across models.]{
        \includegraphics[width=0.3\linewidth]{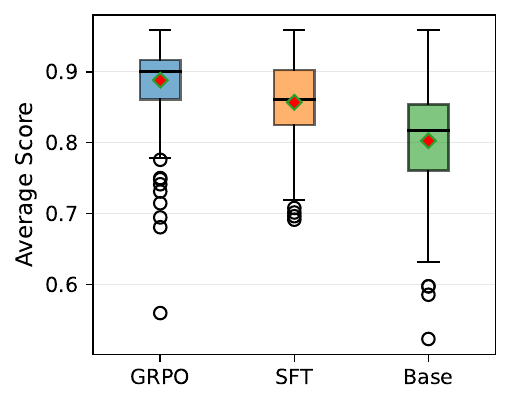}
        \label{fig:outliers}
    }\hfill
    \subfigure[\small Backbone swap ablation.]{
        \includegraphics[width=0.3\linewidth]{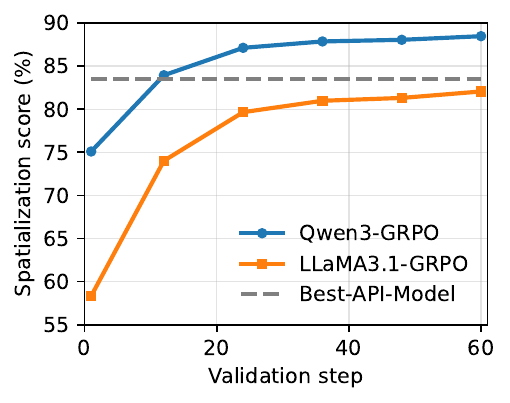}
        \label{fig:backbone_grpo}
    }\hfill
    \subfigure[\small Listener preference predictability.]{
        \includegraphics[width=0.3\linewidth]{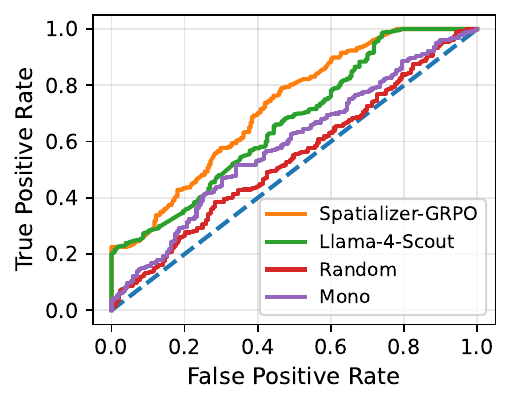}
        \label{fig:roc-curve}
    }

    \caption{\textbf{Additional analyses.} \textbf{(a)} Distribution of per-example AVG scores over GRPO, SFT, and Baseline models. RL training shifts mass and reduces degenerate failures, indicating improved reliability beyond mean score gains.
    \textbf{(b)} Deterministic macro score (\%) versus GRPO steps when swapping the backbone (Qwen3 vs.\ LLaMA~3.3); the horizontal line denotes the best proprietary API baseline (83.45\%).
    \textbf{(c)} ROC curve for a logistic regression that predicts whether listeners prefer audio $A$ over $B$ from the deterministic dramaturgy score difference $\Delta s = s(A)-s(B)$. The classifier achieves $\mathrm{AUC}=0.701$ with Brier score $0.186$.}
    \label{fig:analysis_combined}
    \vspace{-0.24in}
\end{figure*}

\textbf{FanFic Synthetic Augmentation}
To broaden style coverage, we also use the fanfiction collection from \cite{bischoff2020importance}. We process FanFic with the same chunking pipeline and generate \(\sim 5\text{k}\) synthetic training instances. Here the teacher generates \emph{all} annotations (quotes, aliases, positions, movements, and scene metadata) with the same Best-of-64 + rejection procedure. Spatializers are fine-tuned on PDNC + synthetic FanFic, while FanFic is never used for testing.

\subsection{Evaluation Protocol}
\label{sec:exp:eval}

\begin{figure*}[htbp!]
    \centering
    \includegraphics[width=\linewidth]{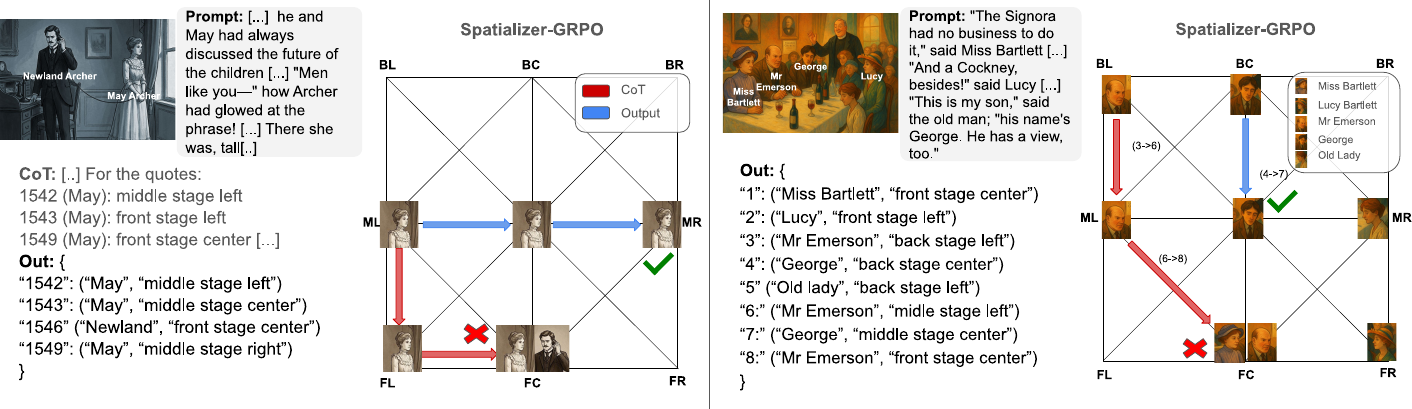}
    \caption{
    \emph{Challenging qualitative cases for spatialization diagnostics.}
    \textbf{Left:} case probing consistency between generated reasoning and emitted layout.
    \textbf{Right:} case probing continuity when new characters enter an already-populated scene.}
    \label{fig:failures}

\vspace{-0.16in}
\end{figure*}

\custompara{Deterministic rule-based scoring.}
We evaluate with our deterministic rule suite, reporting component-wise scores and a macro-average over quotation attribution, alias resolution, stage positioning validity/consistency, character positioning, movement coherence, and scene transitions. We report \textbf{JSON validity} separately; unless noted, scores are on PDNC2 validation.

\custompara{Validating the deterministic rules.}
Our deterministic suite is both a training reward and primary evaluation metric, so we validate it against two external signals: (1) LLM-as-Judge evaluation and (2) a human preference study on spatial audio renderings of predicted layouts:

(1) LLM-as-a-judge: We evaluate outputs with GPT-4.1, which provides (i) an \emph{absolute} rubric-based score in \([0,1]\) (rubric aligned to deterministic dimensions) and (ii) \emph{relative} pairwise preferences aggregated into ELO via Bradley--Terry. We report agreement between deterministic macro-scores and judge absolute scores using Pearson's \(r\). The judge prompt is in \autoref{sec:prompts}.

(2) Human preference study: We convert model outputs into binaural (``spatialized'') audiobook clips and collect pairwise preferences from spatial-audio experts. We also include two negative-control conditions, \textit{mono} and \textit{random}; details 
are in \autoref{sec:userstudy_extra}.

For external validity, we compare evaluator-induced rankings to human win-rate rankings via Spearman correlation ($\rho$), rank accuracy, and Cohen's $\kappa$ between deterministic pairwise predictions
and human pairwise labels.

\custompara{Zero-shot baseline.}
We compare fine-tuned models to prompt-only baselines using the same dramaturgy prompt as in \autoref{sec:method:synthetic-data} (\autoref{sec:prompts}).

\vspace{-0.1in}
\subsection{Implementation Notes}
\label{sec:exp:impl}
\vspace{-0.1in}

Our main backbone is \textsc{Qwen3-8b}~\cite{yang2025qwen3}; we also test \textsc{LLaMA3.1-8b}~\cite{grattafiori2024llama}. Models are trained for 15 epochs with AdamW~\citep{loshchilov2017decoupled} 
in PyTorch/\textsc{verl}~\cite{sheng2024hybridflow}, bfloat16 mixed precision. Training takes \(\sim36\) hours on 64 NVIDIA H100 GPUs; details are in \autoref{sec:app:hyperparameter}.

\section{Results}
\label{sec:results}

\subsection{Deterministic Rule Evaluations}
\label{sec:results:main}

Table~\ref{tab:leaderboard} reports the deterministic suite leaderboard across proprietary and open-source prompt-only baselines, and our trained spatializers. Our Spatializer-GRPO achieves the best overall macro-average (88.5\%), outperforming the strongest open-source baseline (83.4\% AVG; Llama-4-Scout / Llama-3.3-70B-it) by +5.1 points and the best proprietary model (82.0\% AVG; Claude Sonnet 3.5) by +6.5 points. Improvements are concentrated in the global staging components. \textbf{Spatializer-GRPO} achieves the strongest \textbf{Stage Pos.}, \textbf{Char. Pos.}, and \textbf{Move. Coh.} scores in the table, reflecting more reliable blocking (valid placements), tighter stage pictures (better character arrangement), and more coherent motion over time. At the same time, it remains competitive on the local text-anchoring metrics (\textbf{Quote Attr.} and \textbf{Alias Res.}), indicating that these gains do not come from sacrificing attribution or entity resolution.
\begin{table*}[t]
\centering
\small
\renewcommand{\arraystretch}{1.08}
\setlength{\tabcolsep}{6pt}
\begin{tabular}{lccccccc}
\toprule
\textbf{Model} &

\textbf{Quote Attr.} &
\textbf{Alias Res.} &
\textbf{Stage Pos.} &
\textbf{Char. Pos.} &
\textbf{Move. Coh.} &
\textbf{Scene Trans.} &
\textbf{AVG} \\
\midrule
Spatializer-GRPO
&  91.75 & 99.09 & \textbf{92.11} & \textbf{92.54} & \textbf{98.25} & \textbf{57.40} & \textbf{88.52} \\
\midrule
-Only JSON
 & 82.40 & 93.00 & 76.62 & 83.00 & 89.91 & 48.03 & 78.83 \\
-Quote+Alias
 & 27.02 & 31.57 & 51.00 & 72.00 & 75.00 & 39.00 & 49.26 \\
-Stage Pos
 & \textbf{95.19} & 99.23 & 76.00 & 73.74 & 69.00 & 51.07 & 77.37 \\
-Character
 & 94.00 & 99.00 & 66.45 & 57.76 & 65.95 & 49.00 & 72.03 \\
-Movements
 & 93.87 & \textbf{99.37} & 77.74 & 88.02 & 65.00 & 51.07 & 79.18 \\
-Scene
 & 90.53 & 98.33 & 83.19 & 90.34 & 90.46 & 15.05 & 77.98 \\
\bottomrule
\end{tabular}
\vspace{-0.4em}
\caption{Reward-component ablations for the 8B spatializer trained with GRPO. We retrain variants using only JSON validity as reward or by removing one deterministic reward category at a time, and then evaluate using the full deterministic suite. Scores are percentages (higher is better). \textbf{AVG} is the macro-average over Quote Attr., Alias Res., Stage Pos., Char. Pos., Move. Coh., and Scene Trans. (excluding Valid JSON).}
\label{tab:reward_ablations}
\vspace{-1.0em}
\end{table*}
To test whether gains are broad rather than limited to easy examples, Figure \autoref{fig:outliers} shows score distributions for sampled models. Our \textbf{Spatializer-GRPO} produces high-quality spatializations on nearly all prompts with virtually no degenerate failures, whereas strong instruction-tuned baselines exhibit a noticeably heavier tail. Across backbones, we observe the same qualitative shift after RL training: probability mass moves positively, indicating that improvements reflect more consistent spatial reasoning rather than occasional successes.

\subsection{Validating the Deterministic Rule Suite}
\label{sec:validating_dramaturgy_eval}

\begin{table}[t!]
\centering
\small
\renewcommand{\arraystretch}{1.08}
\setlength{\tabcolsep}{4pt}
\begin{tabular}{lcc}
\toprule
\multirow{2}{*}{\textbf{Evaluator}} & \multicolumn{2}{c}{\textbf{Agreement with human pref.}} \\
\cmidrule(lr){2-3}
& \textbf{Spearman} $\boldsymbol{\rho}$ $\uparrow$ & \textbf{Rank Acc.} $\uparrow$ \\
\midrule
LLM-as-Judge (Abs.)          & 0.88 & 86.7 \\
LLM-as-Judge (Rel.)    & 0.86 & 86.7 \\
Deterministic  & \textbf{0.95} & \textbf{93.3} \\
\bottomrule
\end{tabular}
\vspace{-0.35em}
\caption{External validity of automated evaluators against human pairwise preferences on binaural renderings. We report Spearman correlation ($\rho$) between evaluator-induced system rankings and human win-rate rankings, and rank accuracy (Rank Acc.) over all system pairs. Best results are shown in \textbf{bold}.}
\label{tab:human_alignment}
\vspace{-0.9em}
\end{table}

Table~\ref{tab:human_alignment} shows that the deterministic evaluator achieves the strongest agreement with human preferences, with $\rho=0.95$ and $93.3\%$ rank accuracy, outperforming both the absolute and relative LLM-as-Judge variants. This suggests that, despite being hand-designed, the deterministic rules capture a substantial portion of what expert listeners perceive as higher-quality spatialization. 
\begin{table}[t!]
\centering
\small
\setlength{\tabcolsep}{4pt}
\newcommand{\drop}[1]{\textcolor{red}{$^{\downarrow #1}$}}
\newcommand{\grayfont}{\color{gray}}

\begin{tabular}{lcc}
\toprule
\textbf{Model} & \makecell{\textbf{Dramaturgy}\\\textbf{Prompt}} & \makecell{\textbf{Minimal}\\\textbf{Prompt}} \\
\midrule
\multicolumn{3}{l}{\textit{\grayfont Zero-shot models}}\\
\grayfont Llama-4-Scout          & \grayfont 83.39\% & \grayfont 83.25\%\drop{0.14} \\
\grayfont Llama-3.3-70B-Instruct & \grayfont 83.39\% & \grayfont 82.05\%\drop{1.34} \\
\grayfont Llama-4-Maverick       & \grayfont 83.14\% & \grayfont 82.07\%\drop{1.07} \\
\grayfont gemma-3-12b-it         & \grayfont 80.00\% & \grayfont 78.31\%\drop{1.69} \\
\grayfont Qwen3-8b               & \grayfont 78.16\% & \grayfont 77.82\%\drop{0.34} \\
\grayfont Llama-3.3-8B-Instruct  & \grayfont 81.67\% & \grayfont 80.65\%\drop{1.02} \\
\addlinespace
\grayfont \textit{Mean (zero-shot)} & \grayfont 81.63\% & \grayfont 80.69\%\drop{0.93} \\
\midrule
\multicolumn{3}{l}{\textit{Trained models}}\\
Spatializer-SFT           & 84.80\% & 58.33\%\drop{\textbf{26.47}} \\
\rowcolor{blue!5}
\textbf{Spatializer-GRPO} & \textbf{88.50\%} & \textbf{87.82\%}\drop{0.68} \\
\bottomrule
\end{tabular}
\caption{Prompt-sensitivity ablation. We compare the deterministic macro score (AVG) under our full prompt (explicit dramaturgy constraints) versus a minimal prompt that gives only a high-level task description. \textbf{Drop} is computed as (dramaturgy prompt) $-$ (minimal prompt), so larger values indicate stronger dependence on explicit prompt constraints.}
\label{tab:prompt_sensitivity}
\vspace{-0.2in}
\end{table}

The agreement is also reflected at the pairwise level: Cohen's $\kappa$ between deterministic pairwise predictions and human labels is $0.62$, indicating substantial consistency between the deterministic directionality signal and listener preferences. Meanwhile, both judge protocols exhibit high but lower agreement ($\rho\approx 0.86$--$0.88$), consistent with the fact that LLM judges can be informative yet imperfect proxies for perceptual quality.

To further test whether deterministic score differences behave like a meaningful preference signal (rather than only producing a reasonable global ranking), we fit a logistic regression model to predict whether listeners prefer clip $A$ over $B$ from the deterministic score difference $\Delta s = s(A)-s(B)$. Figure~\ref{fig:roc-curve} reports an AUC of $0.701$ with Brier score $0.186$, indicating that $\Delta s$ provides a usable and moderately calibrated predictor of pairwise human preference. Importantly, the negative controls behave as expected: \emph{mono} and \emph{random} spatializations are consistently dispreferred (\autoref{sec:userstudy_extra}), supporting the notion that listeners are not simply rewarding binauralization but are sensitive to coherent, stable staging. Overall, these results suggest that the deterministic evaluation suite is directionally aligned with human judgments and a suitable verifiable reward for rejection fine-tuning and RL post-training.

\section{Ablations}
\label{sec:ablations}

We ablate components of the deterministic dramaturgy reward to identify which signals drive improvements in long-horizon staging. Starting from the same SFT initialization, we retrain the 8B spatializer with GRPO while (i) using only JSON validity as reward, or (ii) removing one reward category at a time (quote attribution + alias resolution, stage positioning validity, character positioning, movement coherence, or scene transitions). All ablated models are evaluated using the \emph{full} deterministic suite (i.e., all checks enabled at evaluation time), so each row reflects how training-time reward shaping affects the full set of dramaturgical criteria.

Table~\ref{tab:reward_ablations} shows that the full reward produces the best overall balance across criteria (88.52 AVG). Using \emph{only} JSON validity as a reward yields mostly well-formed outputs but substantially lower staging quality (78.83 AVG), confirming that syntactic correctness alone does not induce coherent blocking decisions. The most damaging ablation removes the text-grounding terms (``-Quote+Alias''), which collapses attribution and identity consistency and also degrades downstream staging components, indicating that stable \emph{who-spoke-when} tracking is a prerequisite for reliable long-horizon spatial continuity.

The remaining ablations are targeted and interpretable. Removing movement coherence sharply reduces \textbf{Move. Coh.}\ (98.25 $\rightarrow$ 65.00), while removing scene-transition logic collapses Scene Trans.\ (57.40 $\rightarrow$ 15.05) with comparatively smaller changes to within-scene composition. Likewise, removing stage-positioning or character-positioning rewards most strongly harms their corresponding dimensions and lowers the macro-average, supporting the notion that each reward component contributes distinct, non-redundant feedback for improving global staging.

\paragraph{Prompt robustness to dramaturgy constraints}
\label{sec:prompt_robustness}
Our models use a structured ``dramaturgy'' prompt with explicit blocking/composition constraints plus the JSON schema (templates in \autoref{sec:prompts}). We test prompt sensitivity against a \emph{minimal} variant that keeps the same schema but removes most dramaturgical guidance.

Table~\ref{tab:prompt_sensitivity} shows that zero-shot baselines change little when dramaturgy constraints are removed (mean drop: 0.93 points across six models), suggesting prompt engineering alone does not explain performance differences. In contrast, Spatializer-SFT is highly prompt-dependent (84.80\% \(\rightarrow\) 58.33\%), while Spatializer-GRPO remains stable (88.50\% \(\rightarrow\) 87.82\%). This indicates RL from verifiable dramaturgy rewards improves both quality and robustness to prompt simplification.

\paragraph{Backbone ablation.}
To test backbone sensitivity, we replace the GRPO base model from Qwen3 to LLaMA~3.3 while keeping the training recipe fixed. Figure~\ref{fig:backbone_grpo} shows both backbones improve monotonically, but Qwen3 remains consistently stronger (e.g., 75.09\%\(\rightarrow\)88.46\% from step 1 to 60) than LLaMA~3.1 (58.33\%\(\rightarrow\)82.05\%), and exceeds the best proprietary API baseline (83.45\%) after 1 epoch while widening the gap thereafter. This pattern is consistent with recent observations on model-family-specific RLVR behavior \citep{shao2025spuriousrewards,wu2025reasoningormemorization}.

\section{Discussion and Conclusion}
\label{sec:conclusion}
\vspace{-0.1in}

We introduce the \textit{text-to-stage} task: converting narrative prose into \emph{structured} stage-play annotations that specify speaker attribution, entrances/exits, and stage positions for audiobook spatialization. To support learning with objective signals, we combine PDNC quotation/alias labels with pseudo stage directions from a strong teacher, filtered by a deterministic dramaturgy evaluator that uses core staging principles to create a verifiable reward. We then train our Spatializer with SFT followed by GRPO, yielding consistent gains in structural compliance and staging quality 
corroborated by LLM-as-judge signals and human preferences.
 
Looking ahead, the text-to-stage task naturally extends to richer 3D and egocentric staging with explicit listener/camera perspective, as well as continuous movement grammars that model trajectories and durations rather than discrete jumps. Another key direction is multimodal grounding of the layout generation using audio-visual cues---prosody, speaker identity, and sound sources---to better infer intent and emphasis. Finally, stronger mechanisms for long-horizon scene state and 
reasoning-to-action synchrony could further improve scene transitions while preserving the auditability of verifiable rewards.

\section{Limitations}
\label{sec:limitations}
\vspace{-0.1in}

Our method has several limitations. First, performance is still weakest on scene transitions: this remains the lowest-scoring deterministic component and the largest remaining gap relative to stronger baselines, and qualitative diagnostics also show errors in reasoning-to-action consistency 
and entrance continuity, 
especially in long-horizon updates (\autoref{fig:failures}). Second, our representation is intentionally coarse: a discrete \(3\times 3\) stage grid with turn-level position updates cannot capture continuous trajectories, 
which limits physical realism and expressivity. Third, dataset coverage is narrow: evaluation is centered on classic English novels, 
raising potential domain-shift 
concerns for modern dialogue styles, other genres, and non-English settings. Fourth, training and evaluation are tied to the
deterministic reward: though 
we show strong alignment with human preferences, some subjective attributes may still not be captured. 
Fifth, transfer across model families is uneven in our experiments (under the same recipe, Qwen and LLaMA backbones do not benefit equally).

\bibliography{sections/6_references}

\appendix
\counterwithin{figure}{section}
\counterwithin{table}{section}
\renewcommand\thefigure{\thesection.\arabic{figure}}
\renewcommand\thetable{\thesection.\arabic{table}} 

\section*{Appendix}
\section{Deterministic Dramaturgical Reward}
\label{sec:app:deterministic-reward}
To make layout quality verifiable, we design a deterministic rule suite that decomposes the score into the following components:
\begin{itemize}
  \item \textbf{JSON validity} (\(0\)–\(1\) pts): awarded for correctly formatted JSON.
  \item \textbf{Stage position validity} (\(0\)–\(3\) pts): 
    \begin{itemize}
        \item \textbf{Proxemics and depth salience.} Following proxemic zones \citep{hall1966hidden}, we map front/middle/back rows to intimate/personal/social distances on a \(3\times 3\) grid. We reward placing primary interlocutors downstage (front row) and penalize implausible crowding. Let \(d(c)\in\{0,1,2\}\) be the depth row of character \(c\) (0=front). For the top-\(k\) speakers in a scene, we add
        \[
          s_{\text{prox}} \;=\; \frac{1}{k}\sum_{c\in \text{Top-}k} \big(1 - \tfrac{d(c)}{2}\big)\;\in[0,1].
        \]
        \item \textbf{Left--right balance and visual weight.} Inspired by balance and visual weight in composition \citep{mcmanus2011arnheim,block2020visual}, we measure mass asymmetry across the stage.
        Let \(x_t\in\{-1,0,1\}\) be the lateral coordinate at quote \(t\) for all speaking/active characters. The per-scene imbalance is
        \(
         b = \big|\frac{1}{T}\sum_t \mathrm{mean}(|x_t|)\cdot \mathrm{sign}(x_t)\big|.
        \)
        We reward \(b\le \delta\) (default \(\delta=0.4\)) with a margin.
    \end{itemize}
  \item \textbf{Character positioning logic} (\(0\)–\(6\) pts): 
    \begin{itemize}
        \item \textbf{Triangular composition and distinct primaries.} Triangles stabilize focus and avoid colinear staging \citep{block2020visual,hodge2015play}. For the three most active characters with coordinates \((x_i,y_i)\), we compute the normalized triangle area
        \(
         A = \frac{1}{A_{\max}}\cdot \tfrac12 |x_1(y_2-y_3)+x_2(y_3-y_1)+x_3(y_1-y_2)|
        \)
        and give credit if \(A\ge \tau\) and pairwise positions are non-duplicated across consecutive quotes.
    \end{itemize}
    
  \item \textbf{Movement coherence} (\(0\)–\(6\) pts): lateral\({>}\)depth preference; one-zone steps; logical gathering; conflict facing in extended dialogues.
  \begin{itemize}
        \item \custompara{Economy and motivation of movement.} Movement should be legible and scarce without motivation \citep{hodge2015play}. Let \(\Delta_t\) be the Manhattan step between consecutive placements for a character; we reward lateral-first moves and penalize depth oscillations:
        \(
         s_{\text{move}}=\mathrm{mean}_t\big[ \mathbbm{1}\{\Delta_t\le 1\} + \lambda\,\mathbbm{1}\{\text{lateral}\} - \gamma\,\mathbbm{1}\{\text{depth flip}\}\big].
        \)
    \end{itemize}
    
  \item \textbf{Scene transitions} (\(0\)–\(4\) pts): 
  We credit boundaries that coincide with room changes, require new entrants from upstage (back row) unless justified, and cap scene lengths to avoid drift.

  \item \textbf{Quote attribution accuracy} and \textbf{alias resolution accuracy} (both \([0,1]\) pts): The final score sums quote-level micro-credits (attribution correctness; one-cell occupancy; minimal unnecessary motion) and scene-level macro-credits (balance, triangularity, transitions).
\end{itemize}

We aggregate a shaped reward in \([0,1]\):
\[
  r \;=\; \mathbbm{1}_{\text{json}}\cdot
  \frac{\mathrm{QA} + \mathrm{AR} + \frac{\mathrm{SV}}{3} + \frac{\mathrm{CP}}{6} + \frac{\mathrm{MC}}{6} + \frac{\mathrm{ST}}{4}}{6}\,,
\]
where \(\mathbbm{1}_{\text{json}}\!\in\!\{0,1\}\) is the validity gate; \(\mathrm{QA}\), \(\mathrm{AR}\in[0,1]\); and \(\mathrm{SV}\in[0,3]\), \(\mathrm{CP}\in[0,6]\), \(\mathrm{MC}\in[0,6]\), \(\mathrm{ST}\in[0,4]\) are normalized component scores.

\begin{figure*}[ht!]
    \renewcommand{\thefigure}{B.1}
    \centering
    \subfigure[\textbf{Best-of-$N$ selection improves with stronger scorers.}]{
        \includegraphics[width=0.49\linewidth]{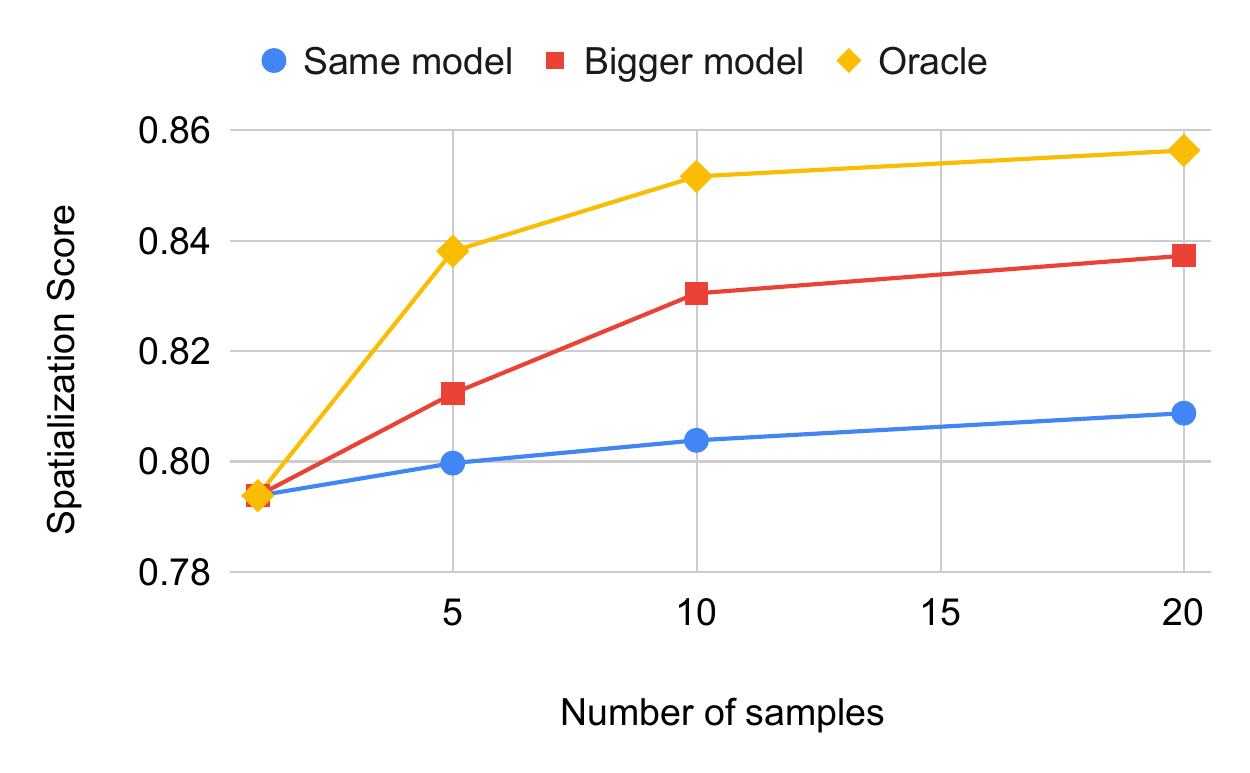}
        \label{fig:best-of-n}
    }\hfill
    \subfigure[\textbf{SFT data scaling shows diminishing returns.}]{
        \includegraphics[width=0.49\linewidth]{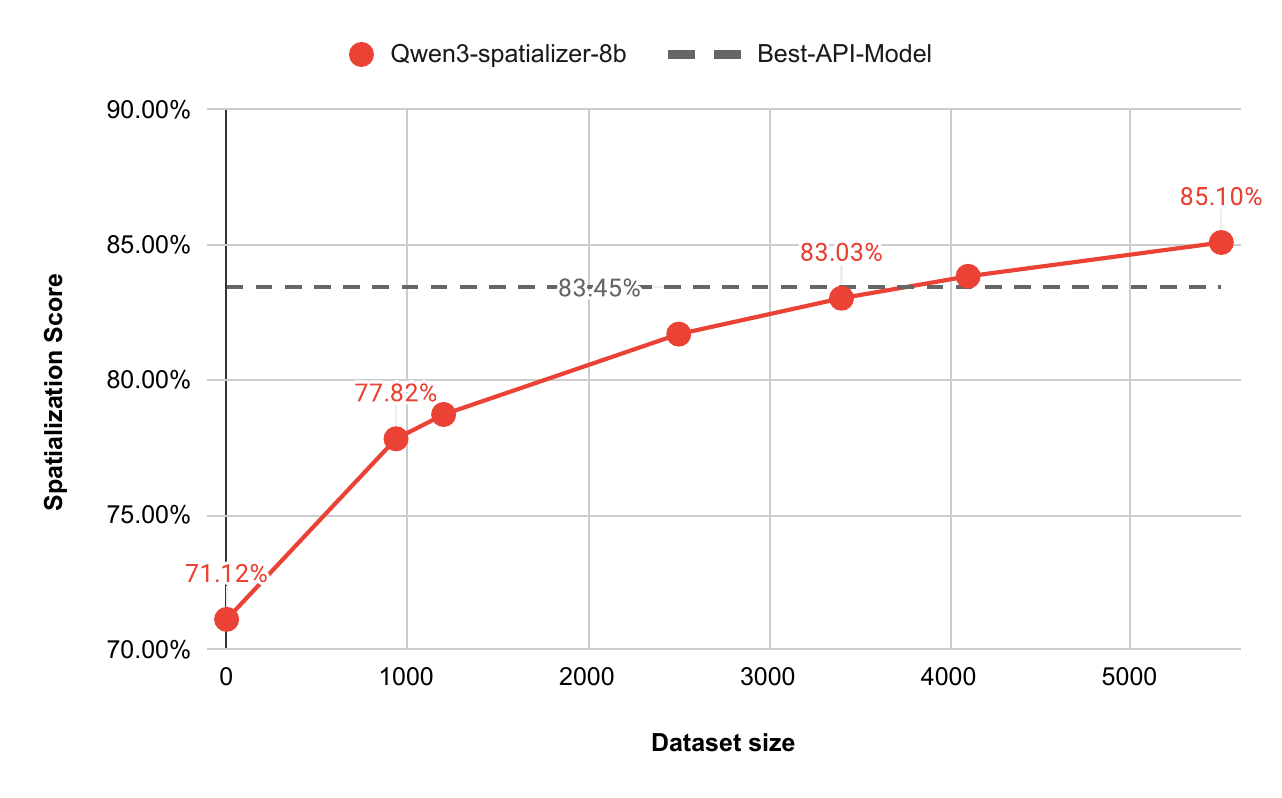}
        \label{fig:scaling}
    }
    \caption{\textbf{Extra results: test-time selection and training-data scaling.}
    \textbf{(a)} Best-of-$N$ at inference time for \texttt{Qwen3-8B}: we sample $N$ candidate spatializations per passage and select the highest-scoring one using three ranking signals---\emph{Same Model} (self-judging with the LLM-as-a-judge prompt), \emph{Bigger Model} (GPT-4.1 as judge with the same prompt), and our deterministic dramaturgy evaluator (\emph{Oracle}). Increasing $N$ improves the selected output for all scorers, with the largest gains under the deterministic oracle.
    \textbf{(b)} Data scaling for \emph{Spatializer-SFT}: we fine-tune the same base model with increasing numbers of supervised training examples and evaluate the resulting model on the same test set. Performance rises monotonically with dataset size, with diminishing marginal gains at larger scales.}
    \label{fig:bestofn-and-scaling}
\end{figure*}
\addtocounter{figure}{-1}

\section{Extra Results}
\label{sec:app:extra-results}
\stepcounter{figure}

\custompara{Best-of-$N$ selection.}
Figure~\ref{fig:bestofn-and-scaling}(a) studies test-time Best-of-$N$ selection for \texttt{Spatializer-SFT} under three scoring signals: (i) \emph{Same Model} self-evaluation with an LLM-as-a-judge prompt, (ii) a stronger judge (\emph{Bigger Model}, GPT-4.1), and (iii) our deterministic dramaturgy evaluator (\emph{Oracle}). Increasing $N$ yields consistent but diminishing returns across all three scorers.
Concretely, from $N{=}1$ to $N{=}20$, the macro score improves from $0.79\!\rightarrow\!0.81$ (+0.02) for self-evaluation, $0.79\!\rightarrow\!0.84$ (+0.05) for GPT-4.1 judging, and $0.79\!\rightarrow\!0.86$ (+0.07) for the deterministic oracle.
Most gains occur early (e.g., Oracle: $0.79\!\rightarrow\!0.84$ already at $N{=}5$), suggesting that a small sample budget captures much of the benefit, while larger $N$ primarily refines selection.
Overall, the deterministic oracle is the most effective selector, indicating that our verifiable reward provides a sharper ranking signal for candidate spatializations than LLM-as-a-judge scoring, while stronger judges (GPT-4.1) bridge part of the gap relative to self-evaluation.

\custompara{Data scaling for Spatializer-SFT.}
Figure~\ref{fig:bestofn-and-scaling}(b) evaluates \texttt{Spatializer-SFT} as we vary the fine-tuning dataset size.
We observe strong improvements from SFT even with modest data: moving from zero data to 937 examples increases the score from $71.12\%\!\rightarrow\!77.82\%$ (+6.70).
Performance continues to rise with additional data, reaching $81.70\%$ at 2{,}500 examples and $85.10\%$ at 5{,}500 examples, but with diminishing marginal gains (e.g., $4100\!\rightarrow\!5500$: +1.26).
These results indicate that (i) the task is highly learnable from relatively small supervised sets, and (ii) additional data still provides measurable benefits, consistent with improved coverage of long-horizon staging patterns and reduced overfitting to narrow narrative styles.

\section{User Study Details}
\label{sec:userstudy_extra}

In order to gauge human preferences regarding model-generated output, we convert the JSON stage-play layout to \textit{spatial audio}, by rendering dialog from an audiobook corresponding to the input text ``spatially" (in front of the listener and preserving positioning and motion cues), at the position determined by the model. We then conducted a standard headphone-based binaural listening test to measure human preference between spatialized audiobook renderings. The study used a 2-interval A/B comparison with no reference: in each trial, assessors listened to two clips (A and B) generated from the same underlying passage by two different models and provided a single overall quality judgment on a 4-level bipolar scale. The scale ranged from 0--3 with the following meanings: 3 = \emph{A is better than B}, 2 = \emph{A is about the same as B}, 1 = \emph{A is worse than B}, and 0 = \emph{both A and B are bad}. Trials spanned 15 unique novels and 45 samples (\texttt{Samp1}--\texttt{Samp45}),
We compare the output from the following models: our \emph{Spatializer-GRPO}, \emph{GPT-4.1}, \emph{Claude Sonnet 3.5}, \emph{Llama-4-Scout-17B-16E-Instruct-FP8}, \emph{Llama-3.3-8B-Instruct}. We additionally include two negative-control systems: \emph{mono} (all dialog is rendered without any spatialization) and \emph{random} (dialogs are assigned to random positions). We had 30 assessors complete 30 pairwise trials each (900 pairwise trials), with randomized source content, model, and placement as option A or B.  To prevent listening fatigue, listeners were encouraged to take a break every 20 mins.

\section{Training Hyperparameters}
\label{sec:app:hyperparameter}
In Table~\ref{tab:spatializer_hyperparams}, we report the detailed dataset usage and training hyperparameters for \textbf{Spatializer-SFT} and \textbf{Spatializer-GRPO}. In summary, \texttt{Stage I} performs supervised fine-tuning (SFT) on \texttt{PDNC-Fanfic} (5{,}500 instruction examples) using rejection fine-tuning, while \texttt{Stage II} applies GRPO on \texttt{PDNC} (800 prompts) using our deterministic dramaturgy evaluator as a verifiable reward. Unless otherwise specified, we train with bfloat16 and PyTorch FSDP on NVIDIA H100 GPUs.

\begin{table}[t!]
\centering
\footnotesize
\resizebox{\linewidth}{!}{
\begin{tabular}{lll}
\toprule[0.95pt]
 & \texttt{Stage I} & \texttt{Stage II} \\
\midrule[0.6pt]
Config & SFT & GRPO \\
\midrule[0.6pt]
\multicolumn{3}{c}{\textit{Training Hyper-Parameters}} \\
\midrule[0.6pt]
Base model & Qwen3-8B & Qwen3-8B \\
Optimizer & AdamW  & AdamW  \\
Learning rate & $5\times 10^{-5}$ & $1\times 10^{-6}$ \\
Training epochs & 1 & 15 \\
Global batch size & 64 & 64 \\
Grad. accumulation & 8  & 1  \\
Context length & 8096  & $5400 \text{ (p)} + 4096 \text{ (r)}$ \\
Rollouts per prompt & \textemdash & 8 \\
KL regularization & \textemdash & $10^{-3}$ \\
\midrule[0.6pt]
\multicolumn{3}{c}{\textit{Training Data}} \\
\midrule[0.6pt]
Dataset & \texttt{PDNC-Fanfic} & \texttt{PDNC} \\
Train set size & 5{,}500 examples & 800 examples \\
Data Type &  Instruction & Reward\\
\midrule[0.6pt]
\multicolumn{3}{c}{\textit{Training Cost}} \\
\midrule[0.6pt]
GPU device & $8\times$ NVIDIA H100 & $64\times$ NVIDIA H100 \\
Compute setup & 1 node $\times$ 8 GPUs & 8 nodes $\times$ 8 GPUs \\
Training time & $\sim$10h & $\sim$36h \\
\bottomrule[0.95pt]
\end{tabular}
}
\caption{\textbf{Training recipes} for \textbf{Spatializer}. \texttt{Stage I}: supervised fine-tuning (SFT) on \texttt{PDNC-Fanfic}. \texttt{Stage II}: GRPO post-training with a deterministic dramaturgy reward on \texttt{PDNC}. $5400 \text{ (p)} + 4096 \text{ (r)}$ means $5400$ are used for the prompt while $4096$ are used for the answer.}
\label{tab:spatializer_hyperparams}
\end{table}

\section{Prompt Templates}
\label{sec:prompts}

This section collects the three prompt templates used throughout the paper. The \emph{Stage-Play Generation} (``dramaturgy'') prompt (Fig.~\ref{fig:prompt-stageplay}) is the primary prompt: given a quoted passage, it (i) attributes each quote to a speaker, (ii) resolves aliases to canonical character names, (iii) assigns discrete 3$\times$3 stage positions under dramaturgical blocking constraints, and (iv) introduces scene splits with lightweight room metadata, all in strict JSON. We use this prompt for prompt-only baselines and as the common input/output format for teacher Best-of-$N$ generation with rejection filtering, as well as during RL. The \emph{Minimal Dramaturgy} prompt (Fig.~\ref{fig:prompt-minimal-dramaturgy}) keeps the same JSON schema but removes most blocking constraints, and is used in the ablation experiments to isolate the effect of detailed dramaturgical guidance. Finally, the \emph{LLM-as-a-Judge} prompt (Fig.~\ref{fig:prompt-judge}) grades candidate JSON outputs with per-criterion integer scores (speaker/alias correctness, stage validity, movement coherence, and scene metadata), and is used as an external evaluator to corroborate our deterministic metrics and ranking trends.

\begin{figure*}[t!]
\centering
    \begin{prompt}{Prompt Template for Stage-Play Generation}
{\color{blue}[Instruction]}
\par
You are an expert dramaturg specializing in adapting literature to stage plays. I will provide you with a passage of a book where quotes have unique identifiers marked by headers ``|quote\_id|''. You are tasked to create a stage play by: (1) sequentially attributing the marked quotes to their speaker, (2) assigning stage positions to characters following dramaturgical principles
\par
\par\relax
{\color{blue}[Inputs]}
\par
{\color{red}[PASSAGE FROM THE BOOK]}
\par
\par\relax
{\color{blue}[Step 1]}
\par
Attribute each quote sequentially to its speaker.
\par
\par\relax
{\color{blue}[Step 2]}
\par
Match each speaker found in the previous step with one of the following names:
\par
\par
{\color{blue}[Names]}
\par
{\color{red}[LIST OF CANONICAL CHARACTER NAMES]}
\par
\par\relax
{\color{blue}[Step 3]}
Replace the speakers found in Step 1 with their matching names found in Step 2, considering:
\par
- How the character is most commonly recognized within the novel \par
- Character relationships and power dynamics \par
- Emotional subtext and narrative importance \par
- Thematic significance of each character's role
\par
\par\relax
{\color{blue}[Step 4]}
Assign a stage position to each character following these dramaturgical principles:
\par
- Use only these 9 positions: \par
\hspace*{1em}\texttt{'back stage left', 'back stage right', 'back stage center',} \par
\hspace*{1em}\texttt{'middle stage left', 'middle stage right', 'middle stage center',} \par
\hspace*{1em}\texttt{'front stage left', 'front stage right', 'front stage center'.} \par
- Back is furthest from the audience, front is closest, and middle is in between. \par
- Position characters based on their narrative importance, objectives, tensions, or alliances evident in the text (dominant speakers downstage (front)/center; minor figures tend upstage (back)). \par
- Character movements must be consistent with the narrative context: \par
\hspace*{1em}$\bullet$ Prefer \textbf{lateral movements} (left to right) instead of depth (front to back); diagonal movements are allowed. \par
\hspace*{1em}$\bullet$ If characters are described as coming together, place them moving toward the same area. \par
\hspace*{1em}$\bullet$ If characters are separating, show them moving away from each other. \par
\hspace*{1em}$\bullet$ If characters are in conflict, position them facing each other. \par
- A character normally moves \textbf{one zone at a time} unless the passage describes a sudden entrance/exit or rapid pursuit. \par
- Keep the overall picture balanced and symmetrical. Asymmetry is allowed when thematically purposeful to highlight imbalance, conflict, or surprise. \par
- If the text implies hierarchy or moral elevation, use depth (back$\rightarrow$middle$\rightarrow$front) or lateral flanks as metaphorical ``high/low ground''. \par
- Reflect power dynamics through proxemics (close proximity signals intimacy or conspiracy; distance signals estrangement). \par
- Ensure \textbf{no two important characters occupy an identical position simultaneously}. Groups or unimportant characters can be clustered together. \par
- Place characters in conversation facing each other (left vs.\ right, or center with either flank). \par
- Position characters to reflect their emotional state (isolation, vulnerability, dominance). \par
- Show character development through gradual position changes (e.g., isolated character moving from upstage to downstage). \par
- New characters typically enter from back positions unless narrative context suggests otherwise (e.g., magical appearances, surprise entrances). \par
- Characters continuing between scenes maintain position if in the same room. \par
- Use movement patterns that maintain narrative flow and visual balance. \par
- An intentionally empty zone (e.g., leaving down-center vacant) can underline absence, danger, or anticipation. \par
- Arrange three or more characters so their positions form a triangle with the audience (isosceles or scalene). Triangles create depth, imply a clear hierarchy, and keep sightlines open.
\par
\par\relax
{\color{blue}[Step 5]}
Split scenes based on these prioritized rules:
\par
1. When characters physically change locations/rooms \par
2. When there's a significant time jump indicated in the narrative \par
3. When the dramatic focus shifts to a new set of characters \par
4. When the emotional tone or dramatic tension significantly changes \par
5. When a conversation has continued for more than 30 quotes \par
6. When the narrator provides extended exposition (3+ sentences) that signals a transition
\par
Never interrupt the natural flow of dialogue unless necessary. Choose scene breaks that enhance dramatic structure and audience comprehension. For each scene, provide:
\par
- Room dimensions (width $\times$ height $\times$ depth) \par
- Wall materials (be specific, e.g., ``oak paneling'' not just ``wood'')
\par
\par\relax
{\color{blue}[JSON Output]}
\par
Your answer should follow this JSON format:
\par
\begin{small}\ttfamily\linespread{0.85}\selectfont
\begin{verbatim}
{
  "scene_1": {
    "room_dimensions": "W x H x D", "room_material": "material",
    "play": {
      "quote_id_1": ["predicted_speaker_1", "stage_position_1"],
      ...
    }
  },
  ...
}
\end{verbatim}
\end{small}
Your answer should only contain the combined outputs of \textbf{Step 3}, \textbf{Step 4}, and \textbf{Step 5} and should strictly adhere to the JSON format. Never generate quote content and don't explain your reasoning.
\end{prompt}
    \caption{{\bf Stage-Play Generation Prompt.} We use the following prompt template to convert a quoted passage into a staged play with canonical speakers, dramaturgically grounded stage positions, and scene splits, outputting structured JSON.}
    \label{fig:prompt-stageplay}
\end{figure*}

\begin{figure*}[t!]
\centering
\begin{prompt}{Minimal Dramaturgy Prompt}
{\color{blue}[Instruction]}
\par
You are an excellent linguist working in the field of literature and play creation. I will provide you with a passage of a book where some quotes have unique identifiers marked by headers ``|quote\_id|''. You are tasked to come up with a stage play: (1) sequentially attribute the marked quotes to their speaker, (2) assign a stage position to characters (one of \texttt{'back stage left', 'back stage right', 'back stage center', 'middle stage left', 'middle stage right', 'middle stage center', 'front stage left', 'front stage right', 'front stage center'}). At the start of each new scene, provide the room dimensions and the wall material.
\par
\par\relax
{\color{blue}[Inputs]}
\par
{\color{red}[PASSAGE FROM THE BOOK]}
\par
\par\relax
{\color{blue}[Step 1]}
\par
Attribute sequentially each quote to its speaker.
\par
\par\relax
{\color{blue}[Step 2]}
\par
Match each speaker found in the previous step with one of the following names:
\par
\par
{\color{blue}[Names]}
\par
{\color{red}[LIST OF CANONICAL CHARACTER NAMES]}
\par
\par\relax
{\color{blue}[Step 3]}
\par
Replace the speakers found in Step 1 with their matching name found in Step 2.
\par
\par\relax
{\color{blue}[Step 4]}
\par
Assign a stage position to each character in the play. Stage positions: \texttt{back} is furthest from the audience, \texttt{front} is closest, \texttt{middle} is in between; \texttt{stage left/right} are from the audience viewpoint. If a character moves, prefer horizontal (left/right) movements over depth (front/back).
\par
\par\relax
{\color{blue}[Step 5]}
\par
At the start of each new scene, provide room dimensions (width $\times$ height $\times$ depth) and describe wall materials. Use common sense to split scenes, but generally cut when the narrator speaks for at least three sentences or characters change rooms.
\par
\par\relax
{\color{blue}[JSON Output]}
\par
Your answer should follow this JSON format (if there is only one scene, still start with \texttt{"scene\_1"}):
\par
\begin{small}\ttfamily\linespread{0.85}\selectfont
\begin{verbatim}
{
  "scene_1": {
    "room_dimensions": "W x H x D",
    "room_material": "material_1",
    "play": {
      "quote_id_1": ["predicted_speaker_1", "stage_position_1"],
      ...
    }
  },
  ...
}
\end{verbatim}
\end{small}
Your answer should only contain the combined outputs of \textbf{Step 3}, \textbf{Step 4}, and \textbf{Step 5} and should strictly adhere to the JSON format. Never generate quote content and don't explain your reasoning.
\end{prompt}
\caption{{\bf Minimal Dramaturgy Prompt.} A simplified prompt template that performs canonical speaker mapping, coarse stage positioning, and scene splitting with minimal dramaturgical constraints, outputting structured JSON.}
\label{fig:prompt-minimal-dramaturgy}
\end{figure*}

\begin{figure*}[t!]
\centering
\begin{prompt}{LLM-as-a-Judge Prompt}
{\color{blue}[Instruction]}
\par
You are an expert dramaturg, stage director, and data-quality auditor.
\par
\par\relax
{\color{blue}[Task]}
\par
You must grade a candidate JSON that assigns speakers and stage positions to the numbered quotes in a passage from a book.
\par
\par\relax
{\color{blue}[Inputs]}
\par
\begin{itemize}\setlength\itemsep{0pt}\setlength\parskip{0pt}
  \item \texttt{PASSAGE}: the excerpt with numbered quotes (do \textbf{not} repeat it in your answer).
  \item \texttt{REFERENCE\_QUOTES}: for every quote-id in the passage, the \textbf{ground-truth speaker}.
  \item \texttt{CANONICAL\_MAPPING}: allowed canonical character names and their aliases.
  \item \texttt{CANDIDATE\_JSON}: the model output you are judging (strict JSON).
\end{itemize}
\par
\par\relax
{\color{blue}[What to Check]}
\par
Evaluate the candidate JSON using the following criteria:
\par
\begin{itemize}\setlength\itemsep{0pt}\setlength\parskip{0pt}
  \item \textbf{A. Quote attribution accuracy} (max 4): Does every quote-id map to the correct canonical speaker (exact match to \texttt{REFERENCE\_QUOTES})?
  \item \textbf{B. Alias resolution} (max 2): Were aliases mapped to the proper canonical form in \texttt{CANONICAL\_MAPPING}?
  \item \textbf{C. Stage position validity} (max 3):
  \begin{itemize}\setlength\itemsep{0pt}\setlength\parskip{0pt}
    \item Does every \texttt{stage\_position} exactly match one of the 9 allowed labels? (1)
    \item Are positions properly distributed across the stage grid? (1)
    \item Is the overall stage picture balanced and symmetrical unless thematically purposeful? (1)
  \end{itemize}
  \item \textbf{D. Character positioning logic} (max 6):
  \begin{itemize}\setlength\itemsep{0pt}\setlength\parskip{0pt}
    \item Dominant speakers downstage (front)/center; minor figures upstage (back). (1)
    \item Characters in conversation face each other (left vs.\ right or center with either flank). (1)
    \item Positions reflect power dynamics, emotional states, and narrative importance. (1)
    \item No two important characters occupy identical positions simultaneously. (1)
    \item With 3+ characters, positions form triangular arrangements when applicable. (1)
    \item Intentionally empty zones used effectively (absence, danger, anticipation). (1)
  \end{itemize}
  \item \textbf{E. Movement coherence} (max 6):
  \begin{itemize}\setlength\itemsep{0pt}\setlength\parskip{0pt}
    \item Prefer lateral (left/right) over depth (front/back) movements. (2)
    \item Typically move one zone at a time unless narrative justifies otherwise. (1)
    \item Characters coming together move toward the same area. (1)
    \item Separating characters move away from each other. (1)
    \item Characters in conflict position themselves facing each other. (1)
  \end{itemize}
  \item \textbf{F. Scene transitions} (max 4):
  \begin{itemize}\setlength\itemsep{0pt}\setlength\parskip{0pt}
    \item Scene boundaries placed appropriately (location/time/focus/tone shifts, $>$30 quotes, or significant exposition). (1)
    \item New characters typically enter from back positions unless context suggests otherwise. (1)
    \item Continuing characters maintain position if in the same room or re-enter appropriately. (1)
    \item Scene changes consistent with room dimension/material changes. (1)
  \end{itemize}
  \item \textbf{G. Metadata quality} (max 3):
  \begin{itemize}\setlength\itemsep{0pt}\setlength\parskip{0pt}
    \item \texttt{room\_dimensions} present and in \texttt{"W x H x D"} format with plausible numbers. (1)
    \item \texttt{room\_material} is specific (e.g., ``oak paneling'' not just ``wood''). (1)
    \item Materials and dimensions match the narrative context. (1)
  \end{itemize}
\end{itemize}
\par
\par\relax
{\color{blue}[Scoring]}
\par
Give an \textbf{integer score} for each criterion (A--G). 0 means the criterion is not met at all. Maximum scores are: A(4), B(2), C(3), D(6), E(6), F(4), G(3). Do \textbf{not} output anything else. Think step-by-step silently, but do \textbf{not} output your reasoning; only output the final JSON.
\par
\par\relax
{\color{blue}[Output Format]}
\par
\begin{small}\ttfamily\linespread{0.85}\selectfont
\begin{verbatim}
{
  "A": <integer score>,
  "B": <integer score>,
  "C": <integer score>,
  "D": <integer score>,
  "E": <integer score>,
  "F": <integer score>,
  "G": <integer score>
}
\end{verbatim}
\end{small}
\end{prompt}
\caption{{\bf LLM-as-a-Judge Prompt.} We use the following prompt to score candidate spatializations (speaker attribution + stage positions + scene metadata) against ground-truth speakers and canonical mappings, producing per-criterion integer scores in JSON.}
\label{fig:prompt-judge}
\end{figure*}

\section{Example Model Trace}
\label{sec:app:model-trace}

Figure~\ref{fig:model-trace} shows one representative model trace for the stage-play generation task. To keep the appendix readable, we shorten the source passage, the intermediate reasoning trace, and the final JSON with \texttt{[...]} wherever content has been omitted without changing the high-level decision pattern.

\begin{sidewaysfigure*}[p]
\centering
\subfigure[\textbf{Input text excerpt.}]{
\begin{minipage}[t]{0.31\textheight}
\vspace{0pt}
\begin{prompt}{Oliver Twist -- Monks Confession: Text}
{\scriptsize\begin{lmttfont}\raggedright
gambled, squandered, forged, and fled to London [...]. She was sinking under a painful and incurable disease, and wished to recover him before she died [...].\\

|4054| ``There she died,'' ||4054|| said Monks, |4055| ``after a lingering illness [...]. I swore to her, if ever it crossed my path, to hunt it down; never to let it rest [...].'' ||4055||\\

[...]\\

|4056| ``The locket and ring?'' ||4056|| said Mr. Brownlow, turning to Monks.\\
|4057| ``I bought them from the man and woman I told you of [...],'' ||4057|| answered Monks [...].\\
|4059| ``Do my eyes deceive me!'' ||4059|| cried Mr. Bumble [...].\\
|4061| ``Hold your tongue, fool,'' ||4061|| murmured Mrs. Bumble.\\
|4065| ``Come, sir,'' ||4065|| said Mr. Grimwig, tartly; |4066| ``suppress your feelings.'' ||4066||\\

[...]\\

|4069| ``Do you know that person?'' ||4069||\\
|4070| ``No,'' ||4070|| replied Mrs. Bumble flatly.\\
|4075| ``You never had, perhaps, a certain gold locket and ring?'' ||4075|| said Mr. Brownlow.\\
|4078| ``Would you like to see the pawnbroker himself?'' ||4078|| asked Mr. Grimwig [...].\\
|4082| ``Nothing,'' ||4082|| replied Mr. Brownlow, |4083| ``except that it remains for us to take care that neither of you is employed in a situation of trust again.'' ||4083||\\

[...]\\

|4102| ``The father of the unhappy Agnes had two daughters,'' ||4102|| said Mr. Brownlow.\\
|4104| ``The child,'' ||4104|| replied Monks, |4105| ``when her father died in a strange place [...], the child was taken by some wretched cottagers [...].'' ||4105||\\
|4108| ``You couldn't find the spot to which these people had repaired,'' ||4108|| said Monks, |4109| ``but where friendship fails, hatred will often force a way.'' ||4109||\\
|4112| ``Do you see her now?'' ||4112||\\
|4113| ``Yes. Leaning on your arm.'' ||4113||\\

[...]\\

|4114| ``But not the less my niece,'' ||4114|| cried Mrs. Maylie [...].\\
|4116| ``The only friend I ever had,'' ||4116|| cried Rose [...].\\
|4120| ``Not aunt,'' ||4120|| cried Oliver [...].\\

[...]\\

|4124| ``I am not here by accident,'' ||4124|| he added [...].\\
|4132| ``The same reasons which influenced me then, will influence me now,'' ||4132|| said Rose firmly [...].\\
|4148| ``I mean but this [...]. This is my rank and station now, and here I lay it down!'' ||4148||
\par\end{lmttfont}}
\end{prompt}
\end{minipage}
}
\hfill
\subfigure[\textbf{Reasoning excerpt.}]{
\begin{minipage}[t]{0.31\textheight}
\vspace{0pt}
\begin{prompt}{Oliver Twist -- Monks Confession: Reasoning}
{\scriptsize\begin{lmttfont}\raggedright
We need to process the passage, assign each quote to its speaker, map aliases to canonical names, assign stage positions, split into scenes, and produce JSON.\\

First identify the main speaking groups: Monks, Mr. Brownlow, Mr. Bumble, Mrs. Bumble, Mr. Grimwig, Rose, Mrs. Maylie, Oliver, and Harry Maylie.\\

[...]\\

Early quote checks:\\
4054 and 4055 are explicit Monks lines.\\
4056 is Brownlow questioning Monks.\\
4059--4060 are Mr. Bumble; 4061 is Mrs. Bumble.\\
4078 is Grimwig introducing corroborating evidence.\\
4102 is Brownlow; 4104--4109 return to Monks's narrative explanation.\\

[...]\\

Alias resolution notes:\\
Map Mrs. Bumble to Mrs. Corney.\\
Map Rose to Rose Maylie and Oliver to Oliver Twist.\\
Map Harry to Harry Maylie.\\
Keep Monks and Mr. Brownlow in their most recognizable canonical forms.\\

[...]\\

Scene split rationale:\\
Scene 1: interrogation in the country house (quotes 4053--4092).\\
Scene 2: Rose's identity is revealed (4093--4121).\\
Scene 3: Harry and Rose in a private exchange (4122--4148).\\
Scene 4: Grimwig's closing remark (4149).\\

[...]\\

Stage positions should reflect dramatic salience: Brownlow stays front stage center during questioning, Monks is held back stage left, Mr. Bumble stays back stage right, Rose is front stage right, and Oliver enters front stage left.\\

More specific placements:\\
Grimwig stays middle stage left as the procedural confirmer.\\
Mrs. Corney stays middle stage right during the accusation sequence.\\
Mrs. Maylie remains back stage right until the family reveal, while Harry dominates front stage center in the final exchange.\\

The final output should therefore group quote IDs by scene and assign each one a canonical speaker plus a stage position.
\par\end{lmttfont}}
\end{prompt}
\end{minipage}
}
\hfill
\subfigure[\textbf{JSON output excerpt.}]{
\begin{minipage}[t]{0.31\textheight}
\vspace{0pt}
\begin{prompt}{Oliver Twist -- Monks Confession: Answer}
{\scriptsize\begin{lmttfont}\raggedright
\{\\
\ \ "scene\_1": \{\\
\ \ \ \ "play": \{\\
\ \ \ \ \ \ "4054": ["Monks", "back stage left"],\\
\ \ \ \ \ \ "4056": ["Mr. Brownlow", "front stage center"],\\
\ \ \ \ \ \ "4059": ["Mr. Bumble", "back stage right"],\\
\ \ \ \ \ \ "4061": ["Mrs. Corney", "middle stage right"],\\
\ \ \ \ \ \ "4065": ["Mr. Grimwig", "middle stage left"],\\
\ \ \ \ \ \ "4078": ["Mr. Grimwig", "middle stage left"],\\
\ \ \ \ \ \ "4082": ["Mr. Brownlow", "front stage center"],\\
\ \ \ \ \ \ [...],\\
\ \ \ \ \ \ "4092": ["Mr. Bumble", "back stage right"]\\
\ \ \ \ \} \\
\ \ \},\\
\ \ "scene\_2": \{\\
\ \ \ \ "play": \{\\
\ \ \ \ \ \ "4093": ["Mr. Brownlow", "front stage center"],\\
\ \ \ \ \ \ "4095": ["Rose Maylie", "front stage right"],\\
\ \ \ \ \ \ "4099": ["Monks", "back stage left"],\\
\ \ \ \ \ \ "4104": ["Monks", "back stage left"],\\
\ \ \ \ \ \ "4112": ["Mr. Brownlow", "front stage center"],\\
\ \ \ \ \ \ "4114": ["Mrs. Maylie", "back stage right"],\\
\ \ \ \ \ \ "4121": ["Oliver Twist", "front stage left"]\\
\ \ \ \ \} \\
\ \ \},\\
\ \ "scene\_3": \{\\
\ \ \ \ "play": \{\\
\ \ \ \ \ \ "4122": ["Harry Maylie", "front stage center"],\\
\ \ \ \ \ \ "4126": ["Rose Maylie", "front stage right"],\\
\ \ \ \ \ \ "4132": ["Rose Maylie", "front stage right"],\\
\ \ \ \ \ \ "4141": ["Harry Maylie", "front stage center"],\\
\ \ \ \ \ \ "4148": ["Harry Maylie", "front stage center"]\\
\ \ \ \ \} \\
\ \ \},\\
\ \ "scene\_4": \{\\
\ \ \ \ "room\_dimensions": "15ft x 12ft x 8ft",\\
\ \ \ \ "room\_material": "brick walls with iron sconces",\\
\ \ \ \ "play": \{\\
\ \ \ \ \ \ "4149": ["Mr. Grimwig", "back stage left"]\\
\ \ \ \ \} \\
\ \ \}\\
\}
\par\end{lmttfont}}
\end{prompt}
\end{minipage}
}
\caption{{\bf Representative model trace for stage-play generation.} A shortened example showing the raw book passage given to the model, an excerpt of the model's intermediate reasoning trace, and the corresponding JSON stage-play output. All three panels are abbreviated with \texttt{[...]} to preserve the most informative decisions while fitting within a single appendix page.}
\label{fig:model-trace}
\end{sidewaysfigure*}

\end{document}